%% file: root.tex
\documentclass[letterpaper, 10 pt, conference]{ieeeconf}  
\usepackage{cite}           
\usepackage[colorlinks=true, citecolor=blue, linkcolor=blue]{hyperref}

\IEEEoverridecommandlockouts                              

\usepackage{amsmath} 
\usepackage{amssymb}  

\usepackage{graphicx}
\usepackage{subcaption}
\usepackage[export]{adjustbox}

\usepackage{tabularx}
\usepackage{multirow}

\usepackage{comment}

\usepackage{booktabs} 
\usepackage[table]{xcolor}

\usepackage{booktabs}
\usepackage{pifont} 
\newcommand{\cmark}{\ding{51}}%
\newcommand{\xmark}{\ding{55}}%

\title{\LARGE \bf
MMD-SLAM: Structure-Enhanced Multi-Meta Gaussian Distribution-Guided Visual SLAM
}

\author{Fan Zhu$^{1,2}$, Ziyu Chen$^{1,2}$, Peichen Liu$^{3}$, Yifan Zhao$^{2}$, Zhisong Xu$^{4}$, \\ 
Hui Zhu$^{1,2}$, Hongxing Zhou$^{5}$, Sixun Liu$^{6}$, and Chunmao Jiang$^{1,*}$
\thanks{*This work was supported by China Postdoctoral Science Foundation under Grant 2025M781668; Anhui Province Key Research and Development Plan under Grant 202423k09020037}
\thanks{$^{1}$ HFIPS, Chinese Academy of Sciences, Hefei, China.}%
\thanks{$^{2}$ University of Science and Technology of China, Hefei, China.}%
\thanks{$^{3}$ Aarhus University, Aarhus, Denmark.}%
\thanks{$^{4}$ University of Tokyo, Tokyo, Japan.}%
\thanks{$^{5}$ Beijing University of Chemical Technology, Beijing, China.}%
\thanks{$^{6}$ North China Electric Power University, Beijing, China.}%
\thanks{$^{*}$ Corresponding author.}
}

\begin{document}

\maketitle
\thispagestyle{empty}
\pagestyle{empty}

\begin{abstract}
3D Gaussian Splatting (3DGS) has significantly boosted novel view synthesis and high-fidelity scene reconstruction, expanding the potential of 3DGS-based Visual Simultaneous Localization and Mapping (SLAM) methods. However, most existing systems fail to fully exploit the underlying structural information, which limits rendering quality and often leads to inconsistent maps. To address these limitations, we propose MMD-SLAM, a structure-enhanced Visual SLAM framework that leverages the Atlanta World (AW) assumption to guide a Multi-Meta Gaussian representation for photorealistic mapping. First, we introduce a point–line fusion strategy for pose optimization, where 3D line segments are incorporated to improve tracking robustness and provide additional constraints for mapping. Second, we design a Multi-Meta Gaussian representation with dominant directions, explicitly encoding structural priors from the AW hypothesis. Finally, we propose a Gaussian evolution strategy that adapts to scene geometry and incorporates structural cues into global optimization. Extensive experiments demonstrate that these innovations enable MMD-SLAM to achieve state-of-the-art performance in both tracking accuracy and mapping quality. e.g., our method achieves a 48.56\% reduction in ATE RMSE on ScanNet and a 5.71\% improvement in PSNR on Replica, compared with MonoGS.
\end{abstract}

\section{Introduction}
\input{txt/01_intro}
\section{Related Works}
\input{txt/02_relatedWork}
\section{Method}
\input{txt/03_method}
\section{Experiment}
\input{txt/04_experiments}
\section{Conclusion}
\input{txt/05_conclusion}

\bibliographystyle{IEEEtranS}
\bibliography{reference}

\end{document}

%% file: txt/01_intro.tex
Visual SLAM aims to estimate camera poses while simultaneously constructing sparse or dense maps of the environment. As a fundamental problem in 3D computer vision, Visual SLAM has been extensively applied in robotics, autonomous driving, and virtual reality (VR)~\cite{focus}.
The traditional Visual SLAM excels at localization, but its limited reconstruction capability falls short for Embodied Artificial Intelligence (Embodied AI). To address this gap, high-fidelity scene reconstruction with new perspective synthesis shows high potential in embodied perception and virtual agent training~\cite{EAI_OV, advance3dv}. Traditional visual SLAM systems primarily focus on camera pose tracking, while the mapping component often relies on sparse representations such as point clouds or voxels. Such low-fidelity maps, which contain holes and lack texture details, are increasingly inadequate for Augmented Reality (AR), VR, and Embodied AI. Consequently, research in Visual SLAM has been shifting toward photorealistic map reconstruction, which is essential for embodied perception and agent training.

The advent of Neural Radiance Fields (NeRF)~\cite{Nerf} has introduced a novel approach to scene representation for robotics~\cite{Ngel-slam}. Integrating NeRF into the Visual SLAM significantly enhances the system's high-fidelity reconstruction capabilities. Despite these advances, NeRF-based methods~\cite{Nice-slam,HS-SLAM} remain limited by over-smoothing and computational inefficiency. 

3D Gaussian Splatting (3DGS)~\cite{3dgs} is an explicit scene expression method that offers substantial improvements in both training speed and rendering quality. However, most 3DGS-based Visual SLAM methods~\cite{Splatam,Photo-slam,MonoGS,Rtg-slam,gs-icp,GARAD-SLAM,FGO-SLAM} explore improvements in rendering quality, efficiency, semantics, and robustness, while ignoring the underlying structural information in the scene. 
With this in mind, MG-SLAM~\cite{MG-SLAM} leverages the Manhattan World assumption and extracts line features for scene optimization. However, its lack of effective integration between discrete features and Gaussian primitives constrains its adaptability.
Furthermore, the anisotropic ellipsoid of 3DGS makes it difficult to reasonably describe these geometric structures, which limits the mapping quality of 3DGS-based methods.

\begin{figure}[!tbp]
	\centering
	\begin{subfigure}[b]{0.49\columnwidth}
		\centering
		\includegraphics[width=\textwidth]{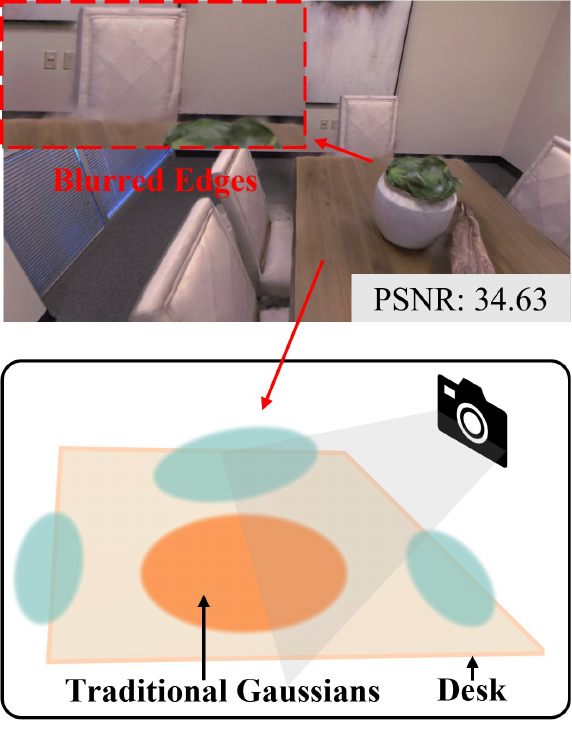}
		\caption{GS-ICP SLAM \cite{gs-icp}}
	\end{subfigure}
	\hfill
	\begin{subfigure}[b]{0.49\columnwidth}
		\centering
		\includegraphics[width=\textwidth]{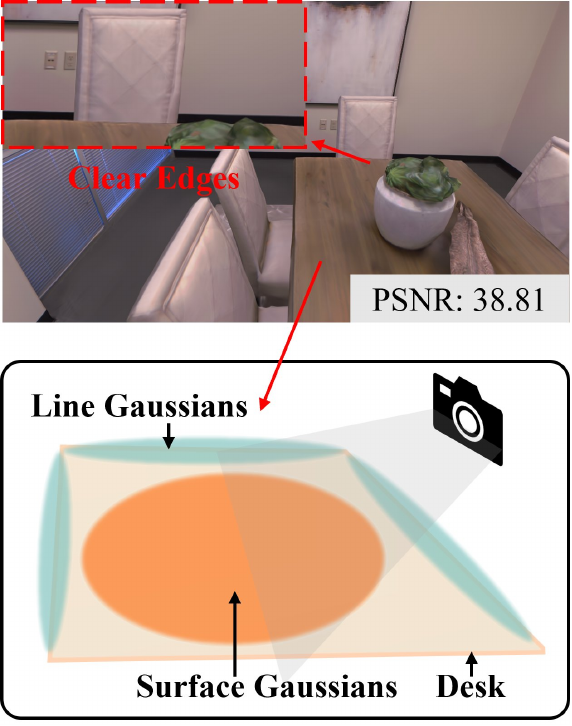}
		\caption{\textbf{Ours}}
	\end{subfigure}
	\caption{\textbf{Comparison of mapping effect.} (a) illustrates that traditional Gaussian ellipsoids, which do not conform to the underlying structure, interfere with each other and generate blurred artifacts. (b) demonstrates that the Multi-Meta Gaussians used in our method better fit the scene structure after training, exhibiting clear edges.}
	\label{fig:abstract}
    \vspace{-6pt}
\end{figure} 

To overcome the above limitations, we proposed MMD-SLAM, a structure-enhanced \textit{\textbf{M}}ulti-\textit{\textbf{M}}eta Gaussian \textit{\textbf{D}}istribution-guided Visual SLAM. Inspired by the Atlanta World (AW) assumption~\cite{AWA}, which postulates that man-made indoor environments are dominated by multiple locally orthogonal directions, we leverage structural priors to improve both tracking and mapping. Specifically, we extract line segment features while extracting feature points in the tracking part, and jointly constrain pose estimation and perform global bundle adjustment optimization by reprojecting point features and back-projecting and reprojecting line features. On the other hand, in the mapping part, based on the AW hypothesis, we divide the 3D Gaussian distribution into three categories: points (spheres), lines, and surfaces, and introduce a dominant direction for this Multi-Meta Gaussian, which is very helpful for the accurate description of geometric structure. At the same time, we propose an evolutionary strategy for Multi-Meta Gaussian, further dividing the Gaussian primitives into Weak states that can flexibly convert types and Stable states that stably represent geometric structures, improving the multi-scene adaptability of our method. Finally, we propose a structure-enhanced scene optimization method. By combining the shape loss and dominant direction loss of Multi-Meta Gaussian on the basis of color and depth loss, we construct a complete loss function, further improving the overall scene reconstruction quality, as shown in Fig. \ref{fig:abstract}.

Overall, the main contributions of our work are as follows:
\begin{enumerate}
    \item We propose MMD-SLAM, a visual SLAM system that leverages the AW hypothesis to guide Multi-Meta Gaussians, leading to significantly improved pose estimation and photorealistic mapping quality. To our knowledge, this innovation is unique.

    \item 
    We propose a structurally enhanced multi-meta representation. By using the AW hypothesis, a set of Multi-Meta Gaussians with dominant directions was innovatively introduced, which can fully utilize the underlying scene structure extracted by the tracking part for mapping. Furthermore, a corresponding optimization strategy are introduced, significantly improving the mapping performance.

    \item Extensive evaluation on both real-world and large-scale synthetic datasets shows that our method achieves state-of-the-art mapping and rendering quality while maintaining competitive tracking accuracy.
\end{enumerate}

%% file: txt/02_relatedWork.tex
\subsection{Classical Dense Visual SLAM}
Over the past decade, extensive work has used visual SLAM for dense scene reconstruction~\cite{Balf}. Traditional research uses explicit representations like point clouds, surfels, and Truncated Signed Distance Functions (TSDF) to reconstruct scenes~\cite{SAGE}. For example, BundleFusion~\cite{Bundlefusion} estimates camera poses in real time from sparse features and uses TSDF for online mapping and map adjustment, achieving globally consistent 3D reconstruction. ElasticFusion~\cite{ElasticFusion} uses surfel sets for scene representation and periodically optimizes the map through non-rigid surface deformation. DI-Fusion~\cite{Di-fusion} integrates scene geometry with deep neural networks for incremental reconstruction. These traditional dense SLAM systems primarily focus on tracking and geometric reconstruction, while our approach also considers high-fidelity appearance reconstruction.
\subsection{Neural Implicit based SLAM}
Following NeRF's rise~\cite{Nerf}, integrating neural implicit representations into SLAM has attracted significant attention. iMAP~\cite{Imap} first demonstrated the potential of implicit representations in SLAM using a single multilayer perceptron (MLP). Subsequently, various works~\cite{Nice-slam,Ngel-slam,HS-SLAM} explored novel implicit scene representations. For instance, NICE-SLAM~\cite{Nice-slam} reconstructs large-scale scenes using hierarchical feature grids with pretrained MLPs. HS-SLAM~\cite{HS-SLAM} combines hash grids, tri-planes, and one-blob networks into a hybrid encoding scheme. Nevertheless, MLP-based implicit methods often suffer from over-smoothing and catastrophic forgetting, limiting their ability to capture fine-grained geometric details.
\subsection{3D Gaussian Splatting based SLAM}
The recent success of 3D Gaussian Splatting (3DGS)~\cite{3dgs} has sparked interest in combining this efficient explicit representation with SLAM systems. Most approaches focus on improving rendering quality~\cite{Splatam,Photo-slam,MonoGS} and computational efficiency~\cite{Rtg-slam,gs-icp}, while others investigate semantic~\cite{Sgs-slam} and robustness~\cite{GARAD-SLAM,SLAM-X} enhancements. However, they generally overlook structural regularities in man-made environments, restricting both rendering quality and mapping consistency. MG-SLAM~\cite{MG-SLAM} addressed this limitation by incorporating structured line features into pose optimization and map refinement, improving reconstruction quality. Yet, it is constrained by the Manhattan World (MW) assumption~\cite{PEMW}, compromising rendering fidelity and adaptability. In contrast, our method generalizes the MW assumption to the Atlanta World (AW) assumption and introduces a set of multi-modal Gaussian primitives that enhance structural modeling while preserving high-quality appearance reconstruction.

%% file: txt/03_method.tex
\begin{figure*}[h]
    \centering
    \includegraphics[width=\linewidth]{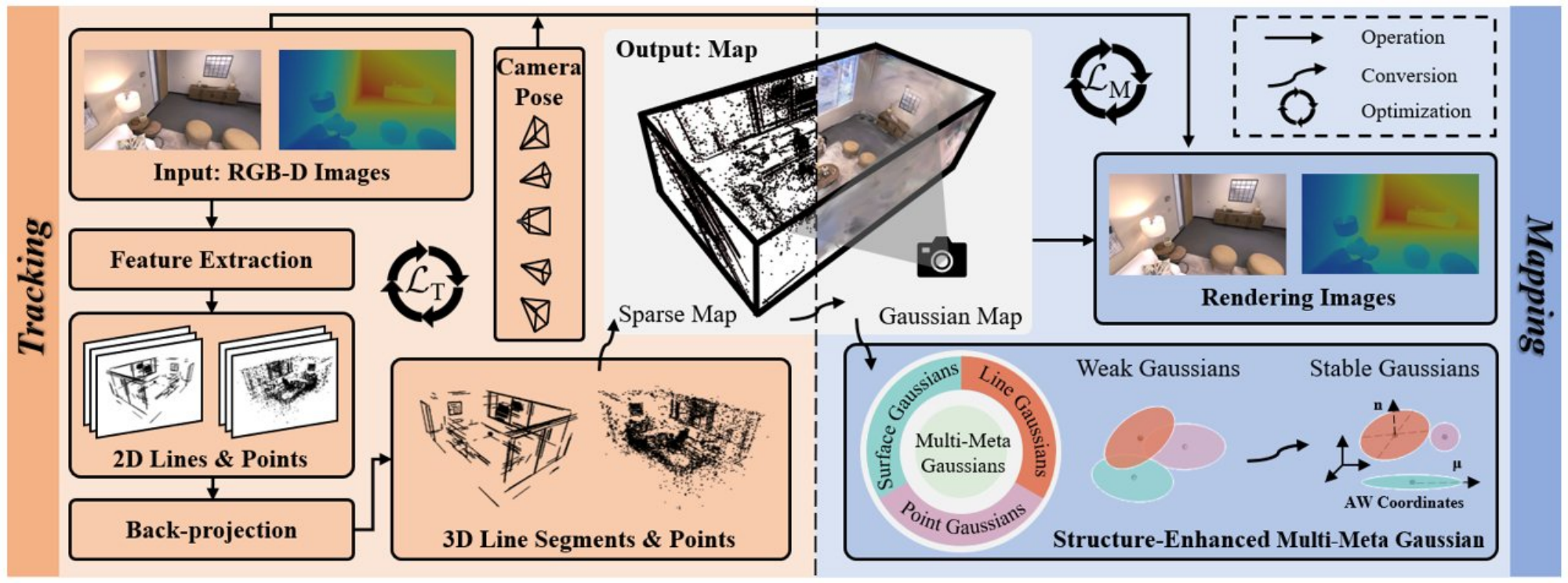}
    \caption{\textbf{Overview.} MMD-SLAM consists of two components: Tracking and Mapping. \textbf{Tracking:} First, extracting point and line features from the input RGB-D frame, the camera pose is determined, and a sparse map is constructed. Secondly, the tracking process is optimized by minimizing the reprojection error and backprojection error. \textbf{Mapping:} Using accurate point cloud information to initialize a Weak Gaussian, a set of stable Multi-Meta Gaussian distributions that conform to the AW assumption is obtained, in order to precisely fit the scene.}
    \label{fig: overview}
    \vspace{-6pt}
\end{figure*}

Fig.~\ref{fig: overview} illustrates the overall pipeline of our approach, which is guided by the Atlanta World (AW) assumption. Section~\ref{Tracking} describes the tracking module, where point–line constraints are incorporated to extract accurate geometric features, construct a sparse map, and perform global optimization, thereby providing reliable structural constraints for subsequent mapping. In Section~\ref{Mapping}, we present the Structure-Enhanced Multi-Meta Gaussian representation as the underlying scene model. By leveraging Gaussian categorization and evolution, we enhance scene reconstruction, and further optimize the map using a combination of appearance and geometry-aware loss functions.

\subsection{Tracking of Point \& Line Fusion Constraints}
\label{Tracking}
\subsubsection{Geometric primitives and notations}
To clearly describe our method, we first introduce the core mathematical notations and geometric primitives.

\textbf{Camera:} Our system is based on the standard pinhole camera model, with intrinsic matrix $\mathbf{K}\in\mathbb{R}^{3\times3}$ and pose $(\mathbf{R},\mathbf{t})\in SE(3)$.

\textbf{Image:} Each 2D image $\mathbf{I}$ is associated with a depth map $\mathbf{D}$. A 2D point and a line segment in the image are denoted as $\boldsymbol{p} = [\mu,\nu]^T\in\mathbb{R}^2$ and $l = (\boldsymbol{p},\boldsymbol{q})$, respectively.

\textbf{3D Space:} In the 3D world domain, given a three-dimensional point $\boldsymbol{P}\in\mathbb{R}^{3\times3}$, we use the projection function $\pi:\mathbb{R}^3\to\mathbf{I}$ to project the three-dimensional point $\boldsymbol{P}_i$ to its corresponding two-dimensional point $\boldsymbol{p}_i$.
At the same time, there is a back-projection function $\beta:\mathbf{I}\to\mathbb{R}^3$ that uses the depth map $\mathbf{D}$ to back-project the 2D pixels to 3D points in the camera coordinate system. A 3D line segment is composed of a pair of 3D points $\left(\boldsymbol{P}, \boldsymbol{Q}\right)\in\mathbb{R}^3\times\mathbb{R}^3$ in the map.
\subsubsection{Point-based pose optimization}
Similar to~\cite{ORB3}, we extract point features from the 2D image and perform back-projection to optimize the camera pose. For the observation point $\boldsymbol{p}_{i}$ in the key frame $\textbf{I}_t$, there is a reprojection error: 
\begin{equation}
\mathcal{L}_{2D}^{p}=\rho\left(\left\|\boldsymbol{p}_{i}-\pi_{k}\left(\mathbf{R}_{k}\boldsymbol{P}_{i}+\mathbf{t}_{k}\right)\right\|_{\mathbf{\Sigma}_{2D}^{p}}\right)
\label{Eq: reprojection error}
\end{equation}
where $\mathbf{\Sigma}_{2D}^p=\alpha_{p_i}^2\mathbb{I}_2$ is the covariance matrix, $\alpha_{p_i}^2\in\mathbb{R}$ is the noise variance of feature points extracted at different scales in the Gaussian image pyramid~\cite{PLVS}, and we denote by $\mathbb{I}_2$ the identity matrix of size $2\times2$.

For RGB-D cameras, we can use depth information to construct a reprojection error of a 3D point. For point $\boldsymbol{p}_{i}$ in the $k$th keyframe, we have 
\begin{equation}
\mathcal{L}_{3D}^p=\rho\left(\left\|\begin{bmatrix}\boldsymbol{p}_i\\\\\boldsymbol{\mu}_i-\frac{bf_x}{d_{p_i}}\end{bmatrix}-\begin{bmatrix}\pi_k\left(\boldsymbol{P}^{W}\right)\\\frac{f_x\left(x^C-b\right)}{z^C}+c_x\end{bmatrix}\right\|_{\mathbf{\Sigma}_{3D}^p}\right)
\label{Eq: RGB-D_reprojection}
\end{equation}
where, the superscripts $W$ and $C$ denote the world and camera coordinate systems, respectively. The variable $d$ represents the depth of point $\boldsymbol{p}_{i}$ in image $\mathbf{I}$; $f_x$ is the focal length in the $x$ direction; $z^C$ is the $z$-coordinate of point $\boldsymbol{P}^C=\mathbf{R}_k\boldsymbol{P}^W+\mathbf{t}_k$ in the camera frame; and $c_x$ is the $x$-coordinate of the principal point.

\subsubsection{Line segment-based pose optimization}
Motivated by~\cite{PLVS}, we incorporate additional line constraints to enhance the robustness of the system. Line segments are efficiently detected at each level of the image pyramid using the EDLines method~\cite{EDLines}. For line matching, we adopt the LBD descriptor~\cite{lbd}, where correspondences are evaluated based on the Hamming distance between descriptors. For a 3D line segment $L\in\mathbb{R}^3$, its projection on the image plane is represented as $l = (\boldsymbol{p}, \boldsymbol{q})$. The back-projection of $l$ is given as $(\beta(\boldsymbol{p}), \beta(\boldsymbol{q}))$. The distance between the camera projection of a 3D point $\boldsymbol{P}^W$ and the 2D line $l$ is denoted as $d^l=\frac{\overline{\boldsymbol{p}}\times\overline{\boldsymbol{q}}}{\left\|\overline{\boldsymbol{p}}\times\overline{\boldsymbol{q}}\right\|}$, where $\overline{\boldsymbol{p}}=[\boldsymbol{p}^T,1]$, $\overline{\boldsymbol{q}}=[\boldsymbol{q}^T,1]$.
Therefore, the 2D line segment reprojection error is defined as:

\begin{equation}
\mathcal{L}_{2D}^l=\rho\left(\left\|\begin{array}{c}d_k^l\cdot\pi\left(\mathbf{R}_k\boldsymbol{P}^W+\mathbf{t}_k\right)\\d_k^l\cdot\pi\left(\mathbf{R}_k\boldsymbol{Q}^W+\mathbf{t}_k\right)\end{array}\right\|_{\mathbf{\Sigma}_{2D}^l}\right)
\label{Eq: 2D_reprojection_error}
\end{equation}

In addition, we exploit the available depth information to overcome the limitations of purely 2D reprojection errors by introducing constraints in 3D space. In the 3D map, the distance between an endpoint $\boldsymbol{P}$ of a line segment and its corresponding back-projected point is defined as: 
\begin{equation}
d_P^l\left(\beta\left(\boldsymbol{p}\right),\boldsymbol{P}^W\right)=\left\|\boldsymbol{P}^C-\beta\left(\boldsymbol{p}\right)\right\|
\end{equation}
where $\boldsymbol{P}^C$ is obtained from $\boldsymbol{P}^C=\mathbf{R}_k\boldsymbol{P}^W+\mathbf{t}_k$. This regularization term constrains the 3D map endpoints around the centroid of their observed back-projected points, preventing drift during optimization.

Furthermore, we measure the perpendicular distance from $\boldsymbol{P}$ to its back-projected 3D line segment:$L^{\prime}=\left(\beta\left(\boldsymbol{p}\right),\beta\left(\boldsymbol{q}\right)\right)$, which enforces alignment between the geometric direction of the 3D line segment and that of its observed back-projection. This yields a 3D line back-projection error:
\begin{equation}
d_{3D}^l\!\left(L^{\prime},\boldsymbol{P}^W\right)
= \scalebox{0.95}{$\displaystyle
    \frac{\left\|\left(\boldsymbol{P}^C-\beta\left(\boldsymbol{p}\right)\right)
    \times
    \left(\boldsymbol{P}^C-\beta\left(\boldsymbol{q}\right)\right)\right\|}
    {\left\|\beta\left(\boldsymbol{p}\right)\times\beta\left(\boldsymbol{q}\right)\right\|}
$}
    \label{Eq: 3D_line_back-projection_error}
\end{equation}
This ensures that the geometric direction of the 3D line segment is aligned with the direction of the currently observed back-projection line segment, from which a back-projection error of the 3D line segment can be obtained:
\begin{equation}
\mathcal{L}_{3D}^l
= \scalebox{0.9}{$\displaystyle
\rho\left(\begin{Vmatrix}d_{3D}^l\left(L^{\prime},\boldsymbol{P}^W\right)+\omega d_P^l\left(\beta\left(\boldsymbol{p}\right),\boldsymbol{P}^W\right)\\d_{3D}^l\left(L^{\prime},\boldsymbol{Q}^W\right)+\omega d_P^l\left(\beta\left(\boldsymbol{q}\right),\boldsymbol{Q}^W\right)\end{Vmatrix}_{\mathbf{\Sigma}_{3D}^l}\right)
$}
\end{equation}
where $\omega\in[0,1]$ is a scalar weighting factor.

\subsubsection{Global BA optimization}
\begin{figure}[t]
    \centering
    \includegraphics[width=1\linewidth]{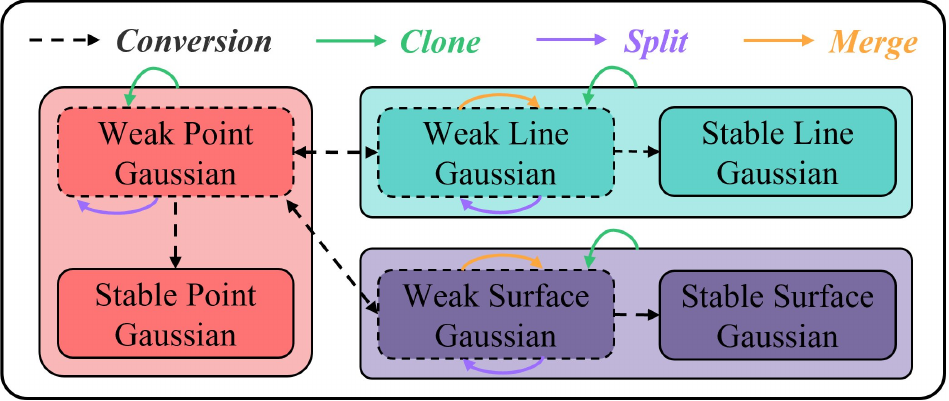}
    \caption{\textbf{Multi-Meta Gaussian Evolution.} The transition from Weak to Stable states is achieved via adaptive type Conversion and density control (Clone, Split, and Merge), a process we term Multi-Meta Gaussian Evolution. Conversion refers to the operation between different types of Gaussians, while densification control is the operation between the same type of Gaussians.}
    \label{fig: Multi-Meta_Gaussian_Evolution}
    \vspace{-6pt}
\end{figure}
Based on the point and line error functions described above, we formulate a global bundle adjustment to minimize the overall objective function as:
\begin{equation}
\mathcal{L}_{T}=\sum_{\mathcal{K}}\bigg(\sum_{\mathcal{P}_{2D}}\mathcal{L}_{2D}^{p}+\sum_{\mathcal{P}_{3D}}\mathcal{L}_{3D}^{p}+\sum_{\mathcal{L}_{2D}}\mathcal{L}_{2D}^{l}+\sum_{\mathcal{L}_{3D}}\mathcal{L}_{3D}^{l}\bigg)
\label{Eq: overall_objective_function}
\end{equation}
 where $\mathcal{K}$ denotes the set of keyframes, and $\mathcal{P}$ and $\mathcal{L}$ represent the sets of points and lines extracted from a given keyframe $k$, respectively.

The optimization is carried out using the Levenberg–Marquardt (L-M) algorithm by iteratively solving the augmented normal as:
\begin{equation}
    \left(\mathbf{J}^T\mathbf{\Sigma}^{-1}\mathbf{J}+\lambda\mathbf{I}\right)\Delta\theta=-\mathbf{J}^T\mathbf{\Sigma}^{-1}\varepsilon
\label{Eq: augmented_normal_equations}
\end{equation}
where the terms $\varepsilon$ and $\mathbf{\Sigma}$ correspond to the projection errors and covariance contributions in the global objective, respectively.

\subsection{Mapping of Structure-Enhanced Multi-Meta Gaussian}
\label{Mapping}
\begin{figure}[t]
    \centering
	\begin{subfigure}[b]{0.48\columnwidth}
		\centering
		\includegraphics[width=\textwidth]{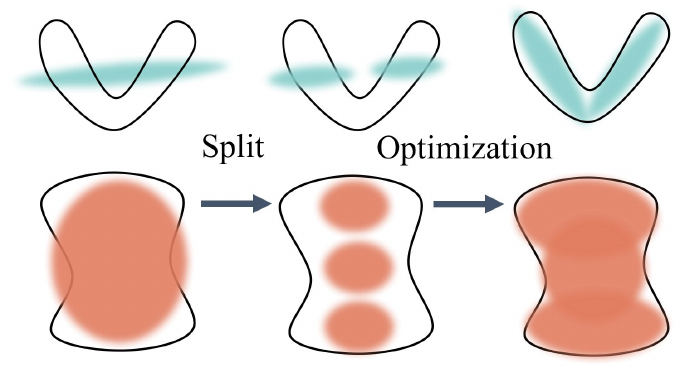}
		\caption{Gaussian Split}
	\end{subfigure}
	\hfill
	\begin{subfigure}[b]{0.48\columnwidth}
		\centering
		\includegraphics[width=\textwidth]{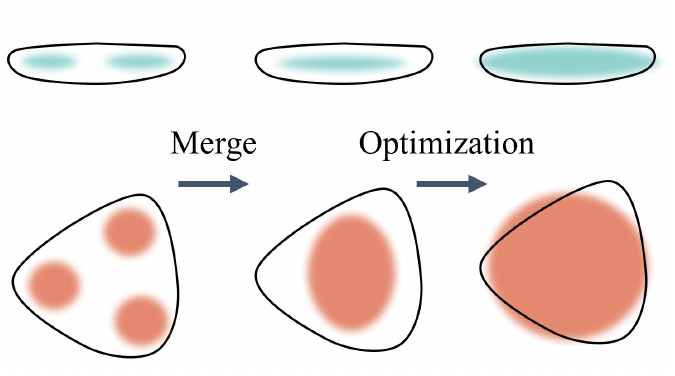}
		\caption{Gaussian Merge}
	\end{subfigure}
    \caption{\textbf{Split and Merge.} We improved the original Split and added Merge to make the Density Control module more suitable for our system.}
    \label{fig: DensityControl}
    \vspace{-6pt}
\end{figure}
\subsubsection{Multi-Meta Gaussian Distribution Guided by AW Assumption}
For 3D Gaussian primitives $\mathcal{G}=\delta N\left(\mu^W,\mathbf{\Sigma}^W\right)$, we largely follow the representation in~\cite{3dgs}. Each Gaussian is parameterized by color $\mathbf{c}$ derived from spherical harmonics $\mathbf{SH}\in\mathbb{R}^{16}$, opacity $\delta\in[0,1]$, mean in world coordinates $\mu^W$, and covariance $\mathbf{\Sigma}^W\in\mathbb{R}^{3\times3}$ decomposed into scale $\mathbf{s}$ and rotation $\mathbf{R}$.

\begin{table*}[!t]
\centering
\caption{\textbf{ATE RMSE (cm) $\downarrow$ results on TUM RGB-D~\cite{tum} and ScanNet~\cite{Scannet} datasets.} The best results are marked as \colorbox{red!40}{best score} and \colorbox{orange!40}{second best score}. The horizontal lines indicate that this method does not support this dataset.}
\label{tab: Tracking}
\begin{tabular}{l|ccc|c|ccccc|c}
\toprule
\textbf{Datasets} 
 & \multicolumn{4}{c|}{TUM RGB-D~\cite{tum}} 
 & \multicolumn{6}{c}{ScanNet~\cite{Scannet}} \\
\cmidrule(lr){2-5} \cmidrule(lr){6-11}
\textbf{Methods} 
 & fr1/desk & fr2/xyz & fr3/office & \textbf{Avg.} 
 & 0000 & 0059 & 0106 & 0169  & 0207 & \textbf{Avg.} \\
\midrule
SplaTAM~\cite{Splatam}   & 3.34 & 1.34 & 5.21 & 3.30 
          & 12.57 & 10.15 & 17.78 & 12.65  & 7.98 & 12.23 \\
MonoGS~\cite{MonoGS}    
          & \cellcolor{red!40}{1.52} & 1.58 & 1.65 & 1.58 
          & 15.94 & 13.03 & 19.44 & 10.44 & 10.46 & 14.20 \\
RTG-SLAM~\cite{Rtg-slam}  
          & 1.62 & \cellcolor{red!40}{0.39} & \cellcolor{orange!40}{1.14} & \cellcolor{red!40}{1.06} 
          & 8.11 & 6.75 & 9.03 & 10.24 & 9.33 & 8.69 \\
GS-ICP SLAM~\cite{gs-icp} 
          & 3.12 & 1.94 & 3.21 & 2.76 
          & -- & -- & -- & --  & -- & -- \\
MG-SLAM~\cite{MG-SLAM}   
          & --   & --   & --   & -- 
          & \cellcolor{orange!40}{5.95} & \cellcolor{red!40}{6.41} & \cellcolor{orange!40}{8.07} & \cellcolor{orange!40}{7.29} & \cellcolor{red!40}{6.14} & \cellcolor{orange!40}{6.77} \\
\textbf{MMD-SLAM (Ours)}      
          & \cellcolor{orange!40}{1.55} & \cellcolor{orange!40}{0.61} & \cellcolor{red!40}{1.12} & \cellcolor{orange!40}{1.09} 
          & \cellcolor{red!40}{5.82} & \cellcolor{orange!40}{6.52} & \cellcolor{red!40}{7.85} & \cellcolor{red!40}{6.99} & \cellcolor{orange!40}{6.51} & \cellcolor{red!40}{6.74} \\
\bottomrule
\end{tabular}
\vspace{-4pt}
\end{table*}

To better fit structured 3D scenes, we draw inspiration from the Atlanta World assumption and extend traditional 3DGS~\cite{3dgs} by introducing three types of Gaussian primitives with distinct geometric modalities: \textbf{\textit{Point Gaussian}} ($\mathcal{G}^P$, \textbf{\textit{Sphere Gaussian}}), \textbf{\textit{Line Gaussian}} ($\mathcal{G}^L$), and \textbf{\textit{Surface Gaussian}} ($\mathcal{G}^S$). Furthermore, each primitive is classified into either a \textit{Stable} or \textit{Weak} state depending on its reliability. For Weak-state Gaussians, we adopt the unified formulation as:
\begin{equation}
\mathcal{G}\left(\mathbf{x}\right)=\exp\left(-\frac{1}{2}\left(\mathbf{x}-\mu^{W}\right)^{T}\left(\mathbf{\Sigma}^{W}\right)^{-1}\left(\mathbf{x}-\mu^{W}\right)\right)
\label{Eq: Weak-state Gaussians}
\end{equation}

For Stable-state Gaussians, the definitions follow three different patterns: Point Gaussian,  Line Gaussian, and Surface Gaussian.

\textbf{Point Gaussian:} Similar to SplaTAM~\cite{Splatam}, we enforce isotropy on Point Gaussians. Their distribution is expressed as:
\begin{equation}
    \mathcal{G}^P\left(\mathbf{x}\right)=\exp\left(-\frac{\left\|\mathbf{x}-\mathbf{\mu}^W\right\|}{2\mathbf{s}^P}\right)
    \label{Eq: Point_Gaussian}
\end{equation}

Comparing with the Eq.~\ref{Eq: Weak-state Gaussians}, Sable Point Gaussian has fewer learned parameters, which can significantly improve training speed.

\textbf{Line Gaussian:} We introduce line Gaussians to model structural edges and other linear features in the scene. For each primitive, among the scale parameters $\mathbf{s}^L=\left[\mathbf{s}_x^L,\mathbf{s}_y^L,\mathbf{s}_z^L\right]$, the largest scale is defined as the \textit{dominant }direction, while the other two scales are close to each other and much smaller. Let $x$ denote the dominant direction, then:
\begin{equation}
\mathbf{s}^L=\left[\mathbf{s}_{dir}^L,\mathbf{s}_{dir}^L/\mathbf{n}^L,\mathbf{s}_{dir}^L/\mathbf{n}^L\right]
\end{equation}

\textbf{Surface Gaussian:} To represent large planar structures in 3D scenes, we define surface Gaussians. In this case, one axis with the smallest scale is defined as the dominant direction, while the remaining two scales are similar and relatively large. This configuration is analogous to the 2DGS representation introduced in~\cite{2dgs}, and its scale characteristics can be written as:
\begin{equation}
\mathbf{s}^S=\left[\mathbf{s}_{dir}^S,\mathbf{s}_{dir}^S\mathbf{n}^S,\mathbf{s}_{dir}^S\mathbf{n}^S\right]
\end{equation}

\subsubsection{Multi-Meta Gaussian Evolution}
Fig.~\ref{fig: Multi-Meta_Gaussian_Evolution} illustrates the operations involved in Multi-Meta Gaussian Evolution. We initialize Gaussian primitives at the 3D points $\boldsymbol{P}$ generated during tracking, with all primitives initially assigned to the Weak state. Through iterative updates, these Weak Gaussians gradually adapt to the surrounding 3D scene structure and are progressively classified into different Multi-Meta Gaussians. 

\textbf{Conversion:} To obtain Stable Gaussians, we reclassify all Weak Gaussians every 1000 iterations and generate new Stable Gaussians. To evaluate this process, we introduce a Type Stability Score (TSS). Specifically, during each classification, if a Weak Gaussian $\mathcal{G}_{W_i}$ retains the same type, its $TSS_i$ increases by 1; otherwise, if a type conversion occurs, $TSS_i$ is reset to 0. When a Gaussian maintains the same type for $n$ consecutive classifications, i.e., $TSS_i = n$, it is promoted to the Stable state, after which its geometric type remains unchanged.

\textbf{Density Control:} Our Clone method largely follows~\cite{3dgs}, with two additional modifications. First, the cloned Gaussian primitive is assigned the same type as the original one. Second, the cloned Gaussian is initialized in the Weak state. These modifications ensure that our approach remains adaptable to scene structures.

Meanwhile, we employ different types of Gaussians for Split and Merge operations to enhance the representation of complex structures, as illustrated in Fig.~\ref{fig: DensityControl}. During the Split process, an oversized Line Gaussian is divided along its dominant vector $u$, splitting the original Gaussian center $\mu^W$ into two centers:
\begin{equation}
\mu_1^W=\mu^W-\frac{1}{2}\phi\mathbf{u},\quad\mu_2^W=\mu^W+\frac{1}{2}\phi\mathbf{u}
\end{equation}
where $\phi$ is a scaling factor. The newly generated Gaussians are set to the Weak state, with their scales also controlled by $\phi$. Similarly, we split the oversized Surface Gaussian into three Weak state Gaussians. The Gaussian centers are distributed triangularly on a plane orthogonal to the principal direction $\mu_i^W=\mu^W+\phi\mathbf{v}_i, i=1,2,3$, where $\mathbf{v}_i$ denotes the in-plane offset vectors. 

To reduce computational overhead, we introduce a Merge operation that fuses neighboring Gaussians of the same type with small scales. Specifically, for merged Line Gaussians, the new dominant scale is defined as the sum of the distances between their centers. For Surface Gaussians, the two in-plane scales are set to the average of the pairwise distances among the three centers. All other attributes are assigned as the mean of the merged Gaussians.

\begin{table*}[!ht]
\centering
\caption{\textbf{Quantitative evaluation of our method compared to SOTA methods on Replica datasets~\cite{replica}.} Colored cells indicate \colorbox{red!40}{best}, \colorbox{orange!40}{second}, and \colorbox{yellow!40}{third} performance. FPS (f/s) represents the operating speed of the entire SLAM system.}
\begin{tabular*}{\textwidth}{@{\extracolsep{\fill}} l l | c c c c c c c c | c | c}
\toprule
\textbf{Methods} & \textbf{Metrics} & Room0 & Room1 & Room2 & Office0 & Office1 & Office2 & Office3 & Office4 & \textbf{Avg.} & FPS $\uparrow$ \\
\midrule
\multirow{3}{*}{SplaTAM~\cite{Splatam}} 
 & PSNR $\uparrow$ & 32.82 & 33.90 & 35.28 & 38.08 & 38.78 & 31.79 & 30.15 & 31.61 & 34.05 & \multirow{3}{*}{} \\
 & SSIM $\uparrow$ & \cellcolor{red!40}0.978 & 0.970 & \cellcolor{red!40}0.984 & 0.981 & 0.982 & \cellcolor{yellow!40}0.965 & 0.951 & 0.946 & 0.970 & 0.12 \\
 & LPIPS $\downarrow$ & \cellcolor{yellow!40}0.072 & 0.096 & \cellcolor{yellow!40}0.070 & 0.089 & 0.095 & \cellcolor{yellow!40}0.098 & 0.118 & 0.156 & 0.099 &  \\
\midrule
\multirow{3}{*}{MonoGS~\cite{MonoGS}} 
 & PSNR $\uparrow$ & \cellcolor{yellow!40}33.91 & \cellcolor{orange!40}35.73 & \cellcolor{orange!40}37.12 & \cellcolor{orange!40}40.28 & \cellcolor{orange!40}41.58 & \cellcolor{yellow!40}35.86 & \cellcolor{orange!40}35.46 & 34.04 & \cellcolor{orange!40}36.75 & \multirow{3}{*}{} \\
 & SSIM $\uparrow$ & 0.946 & 0.957 & 0.964 & 0.972 & 0.976 & 0.962 & 0.958 & 0.941 & 0.960 & 1.59 \\
 & LPIPS $\downarrow$ & \cellcolor{yellow!40}0.072 & \cellcolor{yellow!40}0.077 & \cellcolor{orange!40}0.067 & \cellcolor{yellow!40}0.072 & 0.092 & \cellcolor{yellow!40}0.098 & 0.097 & \cellcolor{yellow!40}0.102 & \cellcolor{yellow!40}0.085 &  \\
\midrule
\multirow{3}{*}{RTG-SLAM~\cite{Rtg-slam}} 
 & PSNR $\uparrow$ & 30.35 & 33.68 & 34.48 & 39.08 & 39.21 & 32.65 & 32.39 & 35.57 & 34.68 & \multirow{3}{*}{} \\
 & SSIM $\uparrow$ & \cellcolor{yellow!40}0.960 & \cellcolor{orange!40}0.977 & \cellcolor{orange!40}0.981 & \cellcolor{red!40}0.989 & \cellcolor{red!40}0.989 & \cellcolor{red!40}0.980 & \cellcolor{orange!40}0.981 & \cellcolor{red!40}0.984 & \cellcolor{orange!40}0.980 &  \colorbox{red!40}{7.14} \\
 & LPIPS $\downarrow$ & 0.156 & 0.115 & 0.122 & 0.082 & 0.102 & 0.143 & 0.138 & 0.123 & 0.123 &  \\
\midrule
\multirow{3}{*}{GS-ICP SLAM~\cite{gs-icp}} 
 & PSNR $\uparrow$ & 32.89 & 35.38 & 34.63 & \cellcolor{yellow!40}40.25 & 40.48 & 34.32 & 34.72 & \cellcolor{orange!40}36.89 & 36.20 & \multirow{3}{*}{} \\
 & SSIM $\uparrow$ & 0.943 & 0.958 & 0.957 & 0.976 & 0.975 & 0.962 & 0.957 & 0.963 & 0.961 & \colorbox{orange!40}{6.85} \\
 & LPIPS $\downarrow$ & 0.075 & \cellcolor{orange!40}0.070 & 0.083 & \cellcolor{orange!40}0.046 & \cellcolor{orange!40}0.056 & \cellcolor{orange!40}0.067 & \cellcolor{orange!40}0.061 & \cellcolor{orange!40}0.063 & \cellcolor{orange!40}0.065 &  \\
\midrule
\multirow{3}{*}{MG-SLAM~\cite{MG-SLAM}} 
 & PSNR $\uparrow$ & \cellcolor{orange!40}34.67 & \cellcolor{yellow!40}35.52 & \cellcolor{yellow!40}37.10 & 40.04 & \cellcolor{yellow!40}41.38 & \cellcolor{orange!40}35.91 & \cellcolor{yellow!40}34.85 & \cellcolor{yellow!40}35.75 & \cellcolor{orange!40}36.90 & \multirow{3}{*}{} \\
 & SSIM $\uparrow$ & \cellcolor{orange!40}0.976 & \cellcolor{red!40}0.978 & \cellcolor{yellow!40}0.980 & \cellcolor{orange!40}0.987 & \cellcolor{orange!40}0.988 & \cellcolor{red!40}0.980 & \cellcolor{orange!40}0.977 & \cellcolor{orange!40}0.978 & \cellcolor{red!40}0.981 & 3.76 \\
 & LPIPS $\downarrow$ & \cellcolor{orange!40}0.070 & 0.084 & \cellcolor{yellow!40}0.070 & 0.076 & \cellcolor{yellow!40}0.083 & 0.101 & \cellcolor{yellow!40}0.095 & 0.112 & 0.086 &  \\
\midrule
\multirow{3}{*}{\textbf{MMD-SLAM (Ours)}} 
 & PSNR $\uparrow$ & \cellcolor{red!40}35.94 & \cellcolor{red!40}38.27 & \cellcolor{red!40}38.81 & \cellcolor{red!40}42.59 & \cellcolor{red!40}41.98 & \cellcolor{red!40}37.51 & \cellcolor{red!40}36.92 & \cellcolor{red!40}38.78 & \cellcolor{red!40}38.85 & \multirow{3}{*}{} \\
 & SSIM $\uparrow$ & 0.958 & \cellcolor{yellow!40}0.971 & 0.964 & \cellcolor{yellow!40}0.982 & \cellcolor{yellow!40}0.983 & \cellcolor{orange!40}0.974 & \cellcolor{yellow!40}0.971 & \cellcolor{yellow!40}0.968 & \cellcolor{yellow!40}0.972 & \colorbox{yellow!40}{3.82} \\
 & LPIPS $\downarrow$ & \cellcolor{red!40}0.031 & \cellcolor{red!40}0.028 & \cellcolor{red!40}0.033 & \cellcolor{red!40}0.018 & \cellcolor{red!40}0.044 & \cellcolor{red!40}0.036 & \cellcolor{red!40}0.027 & \cellcolor{red!40}0.042 & \cellcolor{red!40}0.032 &  \\
\bottomrule
\end{tabular*}
\label{table:rending}
\end{table*}

\begin{figure*}[!ht]
	\centering
	\begin{subfigure}[b]{0.19\linewidth}
		\centering
		\includegraphics[width=\linewidth]{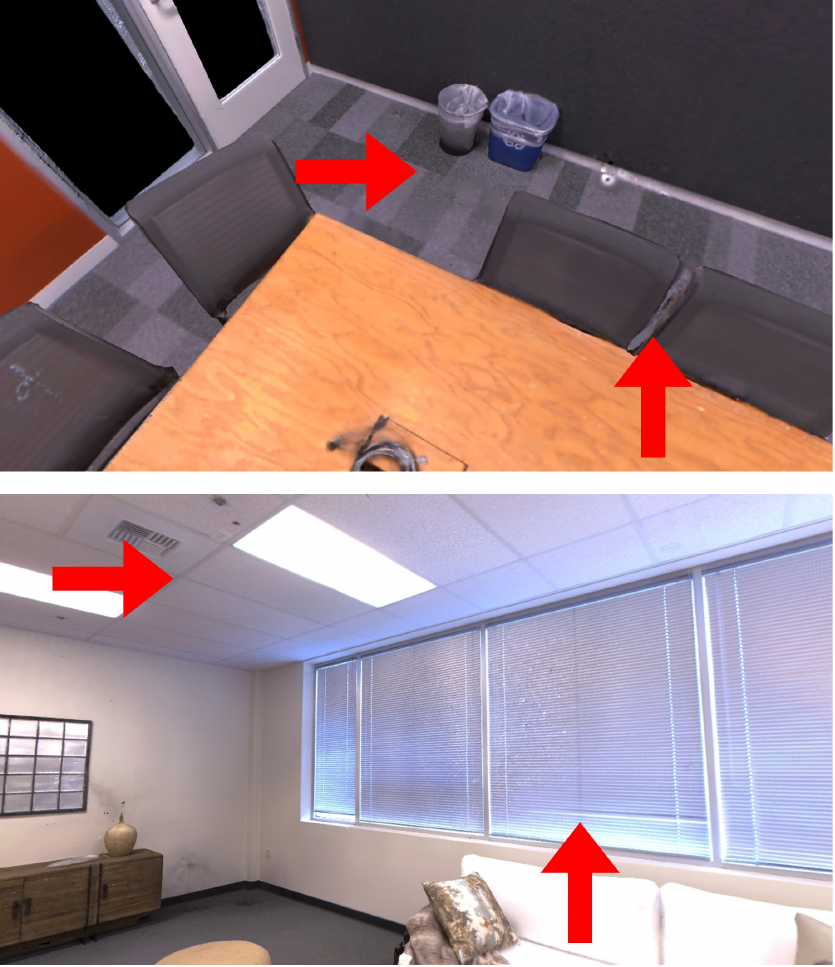}
		\caption{SplaTAM}
	\end{subfigure}
	\hfill
    \begin{subfigure}[b]{0.19\linewidth}
		\centering
		\includegraphics[width=\linewidth]{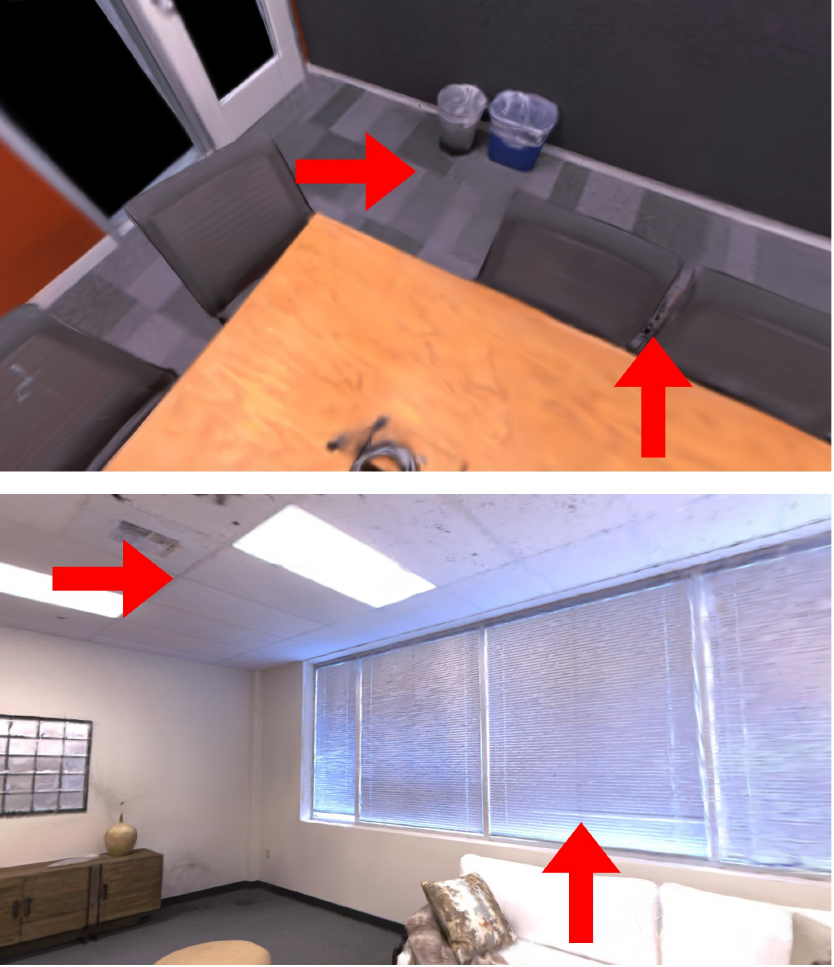}
		\caption{RTG-SLAM}
	\end{subfigure}
	\hfill
	\begin{subfigure}[b]{0.19\linewidth}
		\centering
		\includegraphics[width=\linewidth]{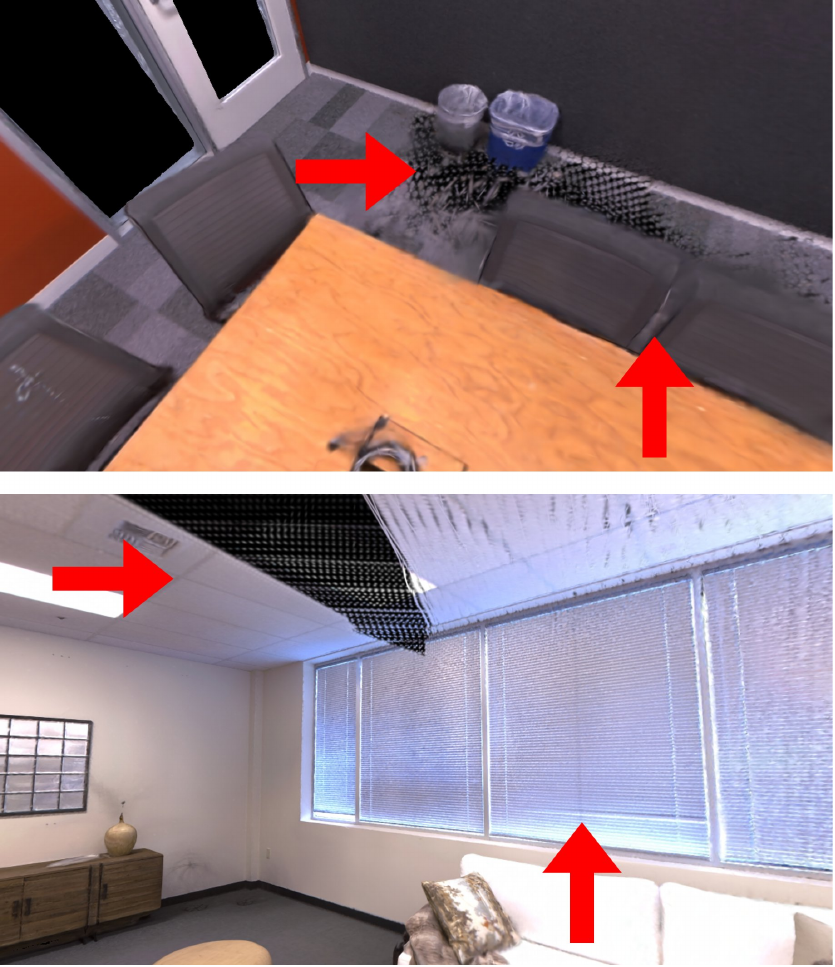}
		\caption{GS-ICP SLAM}
	\end{subfigure}
	\hfill
	\begin{subfigure}[b]{0.19\linewidth}
		\centering
		\includegraphics[width=\linewidth]{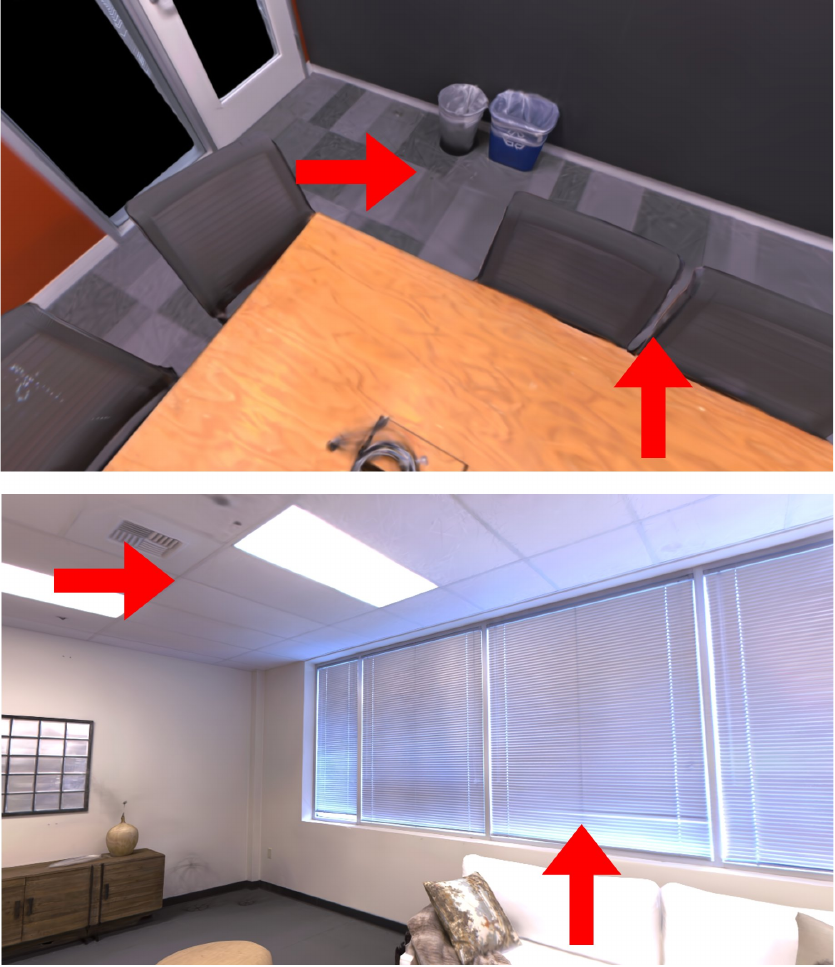}
		\caption{\textbf{Ours}}
	\end{subfigure}
	\hfill
	\begin{subfigure}[b]{0.19\linewidth}
		\centering
		\includegraphics[width=\linewidth]{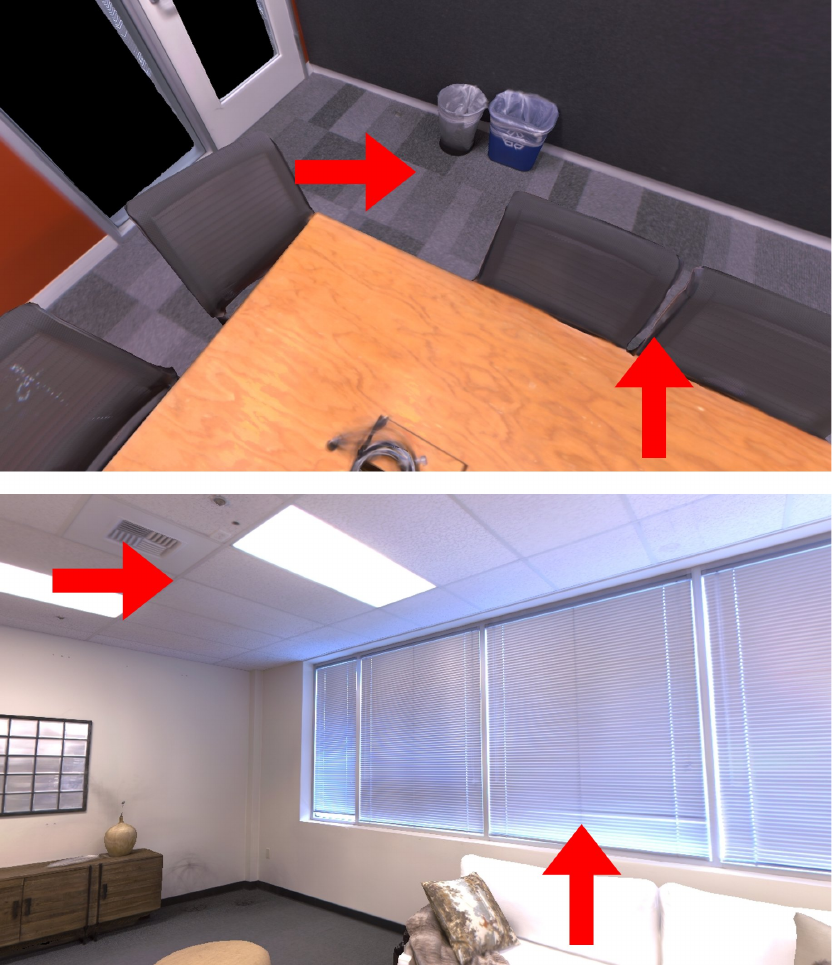}
		\caption{GT}
	\end{subfigure}
	\caption{\textbf{Rendering results on the Replica dataset \cite{replica}.} The red arrow highlight the differences between our approach and baselines. For the details of ceiling, floor and window, our method significantly outperforms baselines.}
	\label{fig:replica}
    \vspace{-6pt}
\end{figure*}

\subsubsection{Structure-enhanced scene optimization}
We introduce a structure-enhanced scene optimization method to optimize the learnable parameters, enabling the Gaussian primitives to fully fit the three-dimensional information. 
Following 3DGS~\cite{3dgs}, we project a 3D Gaussian primitive $\mathcal{G}_i(x)$ onto the 2D image to obtain $\mathcal{G}'_i(x')$. The projected Gaussians are then sorted by depth, and the appearance is rendered using $\alpha$-blending:
\begin{equation}
\mathcal{C}\left(\mathbf{x}^{\prime}\right)=\sum_{i\in N}\mathbf{c}_i\delta_i\mathcal{G}_i^{\prime}\left(\mathbf{x}^{\prime}\right)\prod_{j=1}^{i-1}\left(1-\delta_j\mathcal{G}_i^{\prime}\left(\mathbf{x}^{\prime}\right)\right)
\end{equation}
where $N$ is the number of Gaussians corresponding to the pixel. We use the same method to gain the rendering depth:
$\mathcal{D}\left(\mathbf{x}^{\prime}\right)$. Broadening this into a single image, and we use L1 loss to optimize: 
\begin{equation}
\mathcal{L}_{rgb}=\left\|\mathcal{C}-\mathcal{C}^{GT}\right\|,\quad \mathcal{L}_{depth}=\left\|\mathcal{D}-\mathcal{D}^{GT}\right\| 
\end{equation}

Meanwhile, we introduce a shape constraint loss $\mathcal{L}_{\text{shape}}$. For a single Gaussian primitive $\mathcal{G}_k\left(\mathbf{x}\right)$ from the set of line or surface Gaussians $k \in K$, the scale components along different directions are sorted as $s_{\mathrm{max}}\geq s_{\mathrm{mid}}\geq s_{\mathrm{min}}$. The shape error function is constructed by defining the elongation ratio $e=s_{\max}/s_{\min}$ and flatness ratio $f=s_\mathrm{mid}/s_\mathrm{min}$ of the Gaussian basis:
\begin{equation}
F_L\left(e,f\right)=\frac{1}{e}+\left(f-1\right)^2,\quad F_S\left(e,f\right)=\frac{1}{f}+\left(e-1\right)^2
\end{equation}
To obtain the shape constraint loss:
\begin{equation}
\mathcal{L}_{shape}=\frac{1}{K}\sum_{k=1}^K\left(\rho_k^LF_L\left(e,f\right)+\rho_k^SF_S\left(e,f\right)\right)
\end{equation}

Additionally, using the AW assumption and 3D line segments extracted from the tracking part, we can guide the direction of the line Gaussian:
\begin{equation}
\mathcal{L}_{dir}=\frac{1}{H}\sum_{h=1}^H\left(1-|\mathbf{u}_h\cdot\hat{\mathbf{t}}_k|\right)
\end{equation}
where H is the number of line Gaussians, $\mathbf{u}_h $ is the eigenvector of the dominant direction, and the reference direction provided by the line segment is $\hat{\mathbf{t}}_k$. Consequently, the total loss of this mapping module is
\begin{equation}
\mathcal{L}_M=\mathcal{L}_{rgb}+\lambda_{depth}\mathcal{L}_{depth}+\lambda_{shape}\mathcal{L}_{shape}+\lambda_{dir}\mathcal{L}_{dir}
\end{equation}
where $\lambda_{depth}$  $\lambda_{shape}$, and  $\lambda_{dir}$ are the weight parameters.











%% file: txt/04_experiments.tex
\subsection{Experiment Setup}
\subsubsection{Datasets}
Following~\cite{Nice-slam}, we evaluate our method and baselines on three widely used RGB-D datasets. These include the TUM RGB-D dataset~\cite{tum} for pose evaluation, the larger and more challenging ScanNet dataset~\cite{Scannet}, and the synthetic Replica dataset for mapping evaluation. The Replica dataset version follows the setup of~\cite{Nice-slam}. The selection of the sequence in the ScanNet dataset follows~\cite{MG-SLAM}.
\subsubsection{Baselines}
To evaluate the tracking and mapping performance of our method, we compare against SplaTAM~\cite{Splatam}, MonoGS~\cite{MonoGS}, RTG-SLAM~\cite{Rtg-slam}, GS-ICP SLAM~\cite{gs-icp}, and MG-SLAM~\cite{MG-SLAM}, all of which represent state-of-the-art (SOTA) 3DGS-based SLAM systems.
\subsubsection{Metrics}
We follow the evaluation protocol in~\cite{MonoGS}. For pose estimation, we measure the root mean square error (RMSE) of the absolute trajectory error (ATE) across all frames. For photorealistic mapping, we report standard rendering quality metrics, including PSNR, SSIM, and LPIPS. We calculated the FPS of the entire system for real-time performance.
\subsubsection{Implementation Details}
Our method is implemented on an NVIDIA RTX A100 GPU (40 GB) and an AMD EPYC 7542 CPU. Except for the non-open-source MG-SLAM~\cite{MG-SLAM}, all other baselines are run using their official implementations on the same hardware. To ensure fairness, we report the average results over five runs. For the tracking module, feature extraction parameters follow those used in~\cite{Orbeez-slam}, which is built upon~\cite{ORB3}. For ScanNet, the ground-truth (GT) camera poses are obtained using BundleFusion~\cite{Bundlefusion}. The weight parameters of the mapping process are $\lambda_{_{depth}}=0.1, \lambda_{_{shape}}=0.2, \lambda_{_{dir}}=0.2$.

\begin{figure}[!t]
	\centering
	\begin{subfigure}[b]{0.24\linewidth}
		\centering
		\includegraphics[width=\linewidth]{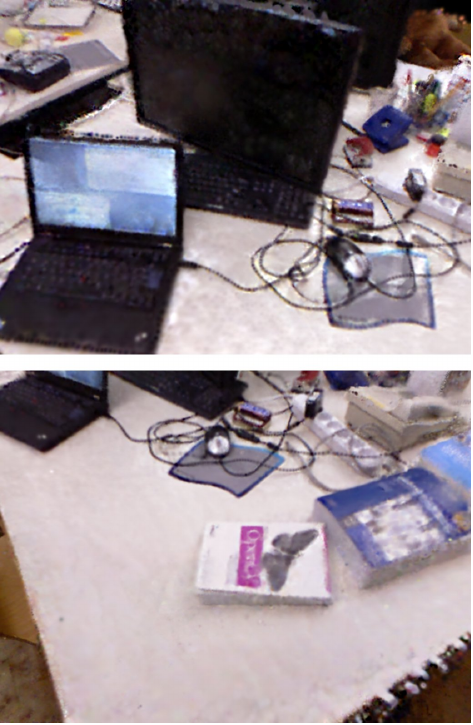}
		\caption{SplaTAM}
	\end{subfigure}
	\hfill
    \begin{subfigure}[b]{0.24\linewidth}
		\centering
		\includegraphics[width=\linewidth]{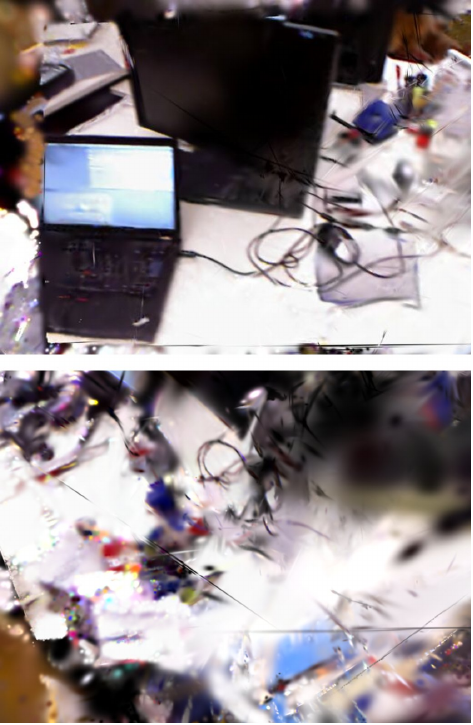}
		\caption{MonoGS}
	\end{subfigure}
	\hfill
	\begin{subfigure}[b]{0.24\linewidth}
		\centering
		\includegraphics[width=\linewidth]{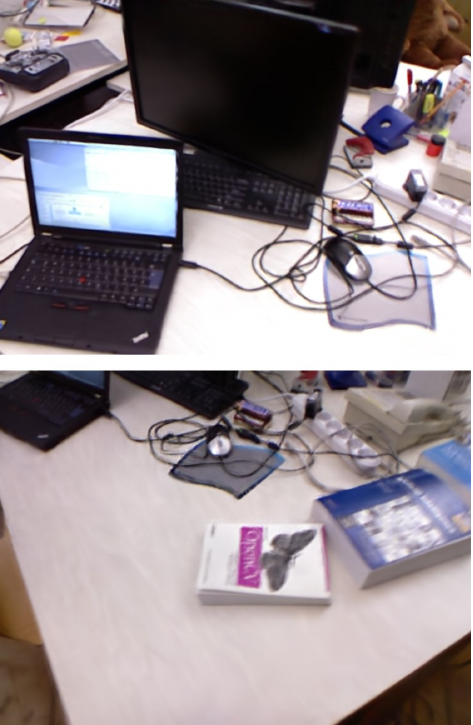}
		\caption{\textbf{Ours}}
	\end{subfigure}
	\hfill
	\begin{subfigure}[b]{0.24\linewidth}
		\centering
		\includegraphics[width=\linewidth]{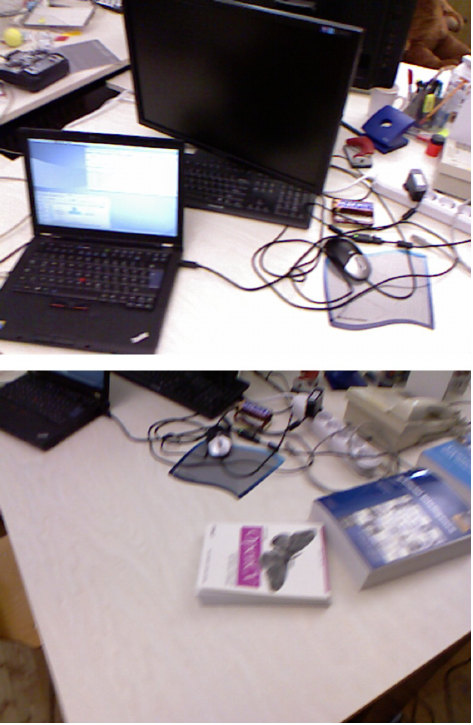}
		\caption{GT}
	\end{subfigure}
	\caption{\textbf{Rendering results on the TUM RGB-D dataset \cite{tum}.} It can be clearly seen that our method can reconstruct a clearer appearance.}
	\label{fig:tum}
    \vspace{-6pt}
\end{figure}

\subsection{Results Analysis}
\subsubsection{Evaluation on Tracking}
As shown in TABLE~\ref{tab: Tracking}, we report comparisons between MMD-SLAM and SOTA baselines on the TUM RGB-D and ScanNet datasets. Our method achieves competitive pose estimation accuracy on both datasets. On the TUM dataset, which contains cluttered structures, ~\cite{Rtg-slam} performs better by relying solely on point features for localization. In contrast, on the larger ScanNet dataset with abundant line features, our method together with~\cite{MG-SLAM} achieves the best results, owing to the similar point–line fusion tracking strategy adopted by both. Overall, when combining the results from both datasets, our method has the best overall positioning performance. This proves the effectiveness of the point-line dual constraint on pose.

\begin{table}[!t]
\centering
\caption{\textbf{Ablation study on the key components.} \colorbox{red!40}{SOTA} results are highlighted. The data is derived from the mean values of eight sequences in the Replica dataset~\cite{replica}.}
\label{tab: ablation_study}
\begin{tabular*}{\linewidth}{@{\extracolsep{\fill}}ccc|cc}
\toprule
PLFC & MMGE & SSO & ATE RMSE (cm) $\downarrow$ & PSNR (dB) $\uparrow$ \\
\midrule
\xmark & \xmark & \xmark & 0.61 & 35.92 \\
\cmark & \xmark & \xmark & \cellcolor{red!40}{0.47} & 37.78 \\
\cmark & \cmark & \xmark & \cellcolor{red!40}{0.47} & 38.31 \\
\cmark & \cmark & \cmark & \cellcolor{red!40}{0.47} & \cellcolor{red!40}{38.85} \\
\bottomrule
\end{tabular*}
\vspace{-6pt}
\end{table}

\subsubsection{Evaluation on Mapping}
Since the original image quality of the TUM RGB-D and ScanNet datasets is relatively low, we perform quantitative mapping evaluation only on the Replica dataset. TABLE~\ref{table:rending} reports the mapping results of our method compared with baselines, where MMD-SLAM significantly outperforms all competitors and has a competitive running speed. Moreover, to better demonstrate reconstruction quality, we conduct qualitative comparisons on the TUM RGB-D and Replica datasets, as shown in Fig.~\ref{fig:tum} and Fig.~\ref{fig:replica}. In particular, the comparison on the TUM RGB-D dataset (Fig.~\ref{fig:tum}) clearly shows that our method achieves the best rendering quality. Finally, as illustrated in Fig.~\ref{fig:replica}, SplaTAM, which relies entirely on isotropic Gaussians, often produces holes in the map; RTG-SLAM suffers from generally poor rendering quality; and GS-ICP SLAM exhibits significant defects in reconstructed floors and ceilings. In contrast, we employed the Multi-Meta Gaussian model guided by the AW hypothesis, fully leveraging the structural information and achieving the best detail restoration.

\subsubsection{Ablation Study}
In order to demonstrate the role of key components in our system, we provide a comprehensive ablation analysis to demonstrate their importance.

\textbf{Point \& Line Fusion Constraints (PLFC):} The first two rows of TABLE~\ref{tab: ablation_study} present the ablation study on the tracking module with respect to Point \& Line Fusion Constraints (PLFC). Row 1 corresponds to tracking using only point features, while Row 2 incorporates both point features and line segments. The results show that introducing additional line constraints plays a critical role in pose optimization and contributes to improved mapping quality.

\begin{figure}[!t]
	\centering
	\begin{subfigure}[b]{0.49\linewidth}
		\centering
		\includegraphics[width=\linewidth]{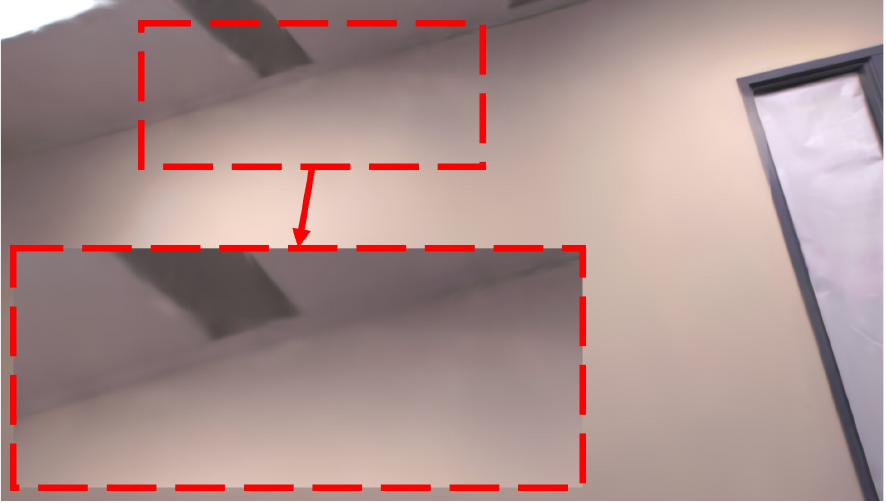}
		\caption{w/o MMGE \& SSO}
	\end{subfigure}
	\hfill
	\begin{subfigure}[b]{0.49\linewidth}
		\centering
		\includegraphics[width=\linewidth]{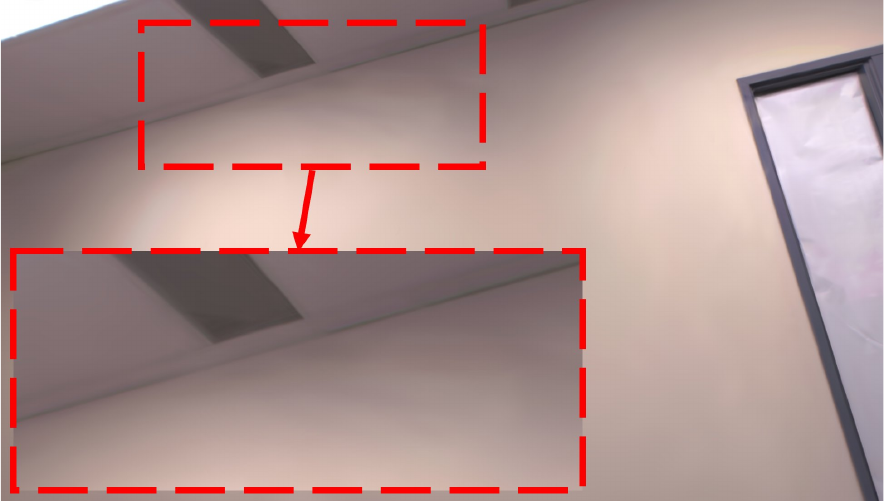}
		\caption{w MMGE \& SSO}
	\end{subfigure}
	\caption{\textbf{Ablation of MMGE \& SSO.} The red dashed box highlights the differences between the two methods. Our comprehensive approach has more obvious structure.}
	\label{fig:ablation}
    \vspace{-4pt}
\end{figure}

\textbf{Multi-Meta Gaussian Evolution (MMGE):} To validate the effectiveness of the Multi-Meta Gaussian Evolution (MMGE) component in MMD-SLAM, we train a standard 3DGS model using the same poses. Row 2 reports the mapping results with single-modal Gaussian primitives, while Row 3 shows the results with our Multi-Meta Gaussian primitives under the same poses. This comparison demonstrates the effectiveness of the proposed Multi-Meta Gaussians.

\textbf{Structure-Enhanced Scene Optimization (SSO):} We combine appearance and geometric losses to design a Structure-Enhanced Scene Optimization (SSO) method. We conduct an ablation by controlling whether geometric losses are included. Row 4 in TABLE~\ref{tab: ablation_study} corresponds to our full method, showing that each component contributes critically to the overall performance. Fig.~\ref{fig:ablation} visualizes the effects of the key components of our method.

%% file: txt/05_conclusion.tex
In this work, we presented MMD-SLAM, a novel 3DGS based SLAM framework. Our method leverages point and line features as complementary constraints for robust pose estimation and mapping guidance. Guided by the Atlanta World hypothesis, we construct a set of Multi-Meta Gaussian distributions as the underlying scene representation, incorporating Gaussian evolution strategies and structure-enhanced optimization to accurately capture both appearance and geometry. Extensive experiments demonstrate that MMD-SLAM achieves SOTA performance. In future work, we plan to extend the applicability of our approach and further enhance its robustness in challenging environments.